\documentclass[12pt,letterpaper]{article}
\usepackage[a4paper, total={7in, 10in}]{geometry}

\usepackage{graphicx}
\usepackage{helvet}
\usepackage{authblk}
\usepackage{hyperref}
\usepackage{amsmath} 
\usepackage{amssymb} 
\usepackage{orcidlink} 
\usepackage[super,comma,sort&compress]  
   {natbib}
\usepackage[right]{lineno} \linenumbers
\usepackage[title]{appendix}%

\usepackage[utf8]{inputenc} 
\usepackage[T1]{fontenc}    
\usepackage{url}            
\usepackage{booktabs}       
\usepackage{amsfonts}       
\usepackage{nicefrac}       
\usepackage{microtype}      
\usepackage{xcolor}         
\usepackage{multirow}
\usepackage{threeparttable}
\usepackage{subcaption}
\usepackage{placeins}
\usepackage{siunitx}       
\usepackage{caption}       
\usepackage{tabularx}
\usepackage{listings}    
\usepackage{pifont}
\usepackage{array}
\usepackage{makecell}
\theadset{\bfseries}
\usepackage[most]{tcolorbox}
\usepackage{rotating}

\newcommand{\mhead}[1]{\makecell[c]{#1}}

\makeatletter
\renewcommand{\maketitle}{\bgroup\setlength{\parindent}{0pt}
\begin{flushleft}
  \textbf{\@title}
  
  \@author
\end{flushleft}\egroup}
\makeatother

\title{mtslearn: Machine Learning in Python for Medical Time Series}
\date{}

\author[1,2]{Zhongheng Jiang}
\author[2,3]{Yuechao Zhao}
\author[2,4]{Donglin Xie}
\author[5,6]{Chenxi Sun}
\author[7]{Rongchen Lu}
\author[8]{Silu Luo}
\author[2]{Zisheng Liang}
\author[2,4,9,10,*,\orcidlink{0000-0001-7521-5127}]{Shenda Hong}

\affil[1]{Nanjing University of Information Science and Technology, Nanjing, China}
\affil[2]{National Institute of Health Data Science, Peking University, Beijing, China}
\affil[3]{School of Health Humanities, Peking University Health Science Center, Beijing, China}
\affil[4]{Institute of Medical Technology, Peking University Health Science Center, Beijing, China}
\affil[5]{Harvard Medical School, Boston, US}
\affil[6]{Beth Israel Deaconess Medical Center, Boston, US}
\affil[7]{School of Business English, Sichuan International Studies University, Chongqing, China}
\affil[8]{School of Clinical Medicine, Chengdu Medical College, Chengdu, Sichuan, China}
\affil[9]{Institute for Artificial Intelligence, Peking University, Beijing, China}
\affil[10]{State Key Laboratory of Vascular Homeostasis and Remodeling, NHC Key Laboratory of Cardiovascular Molecular Biology and Regulatory Peptides, Peking University, Beijing, China}

\affil[*]{Correspondence: hongshenda@pku.edu.cn}

\newtcolorbox[auto counter]{mybox}[2]{
    colback=gray!5,
    colframe=gray!80,
    arc=0pt,
    breakable,          
    enhanced,           
    title={listing \thetcbcounter: #2},
    label={#1}
}

\begin{document}

\maketitle

\section*{ABSTRACT}
Medical time-series data captures the dynamic progression of patient conditions, playing a vital role in modern clinical decision support systems. However, real-world clinical data is highly heterogeneous and inconsistently formatted. Furthermore, existing machine learning tools often have steep learning curves and fragmented workflows. Consequently, a significant gap remains between cutting-edge AI technologies and clinical application. To address this, we introduce mtslearn, an end-to-end integrated toolkit specifically designed for medical time-series data. First, the framework provides a unified data interface that automates the parsing and alignment of wide, long, and flat data formats. This design significantly reduces data cleaning overhead. Building on this, mtslearn provides a complete pipeline from data reading and feature engineering to model training and result visualization. Furthermore, it offers flexible interfaces for custom algorithms. Through a modular design, mtslearn simplifies complex data engineering tasks into a few lines of code. This significantly lowers the barrier to entry for clinicians with limited programming experience, empowering them to focus more on exploring medical hypotheses and accelerating the translation of advanced algorithms into real-world clinical practice. mtslearn is publicly available at \url{https://github.com/PKUDigitalHealth/mtslearn}.

\section*{KEYWORDS}
Medical Time Series, Machine Learning Toolkit, Python, Clinical Decision Support Systems

\section*{INTRODUCTION}
With the widespread adoption of electronic health records \cite{shen2025} and modern medical monitoring devices \cite{luo2024}, medical data is experiencing significant expansion. Within this massive volume of data, the vast majority, such as vital signs and medication records, are fundamentally time-series \cite{saveliev2023}. These time-series data capture the dynamic evolution of patient states. By leveraging effective machine learning techniques, they can systematically support clinical decision-making \cite{niu2025}.

Despite the proven efficacy of machine learning in healthcare and the proliferation of machine learning tools, a significant divide persists between artificial intelligence technologies and clinical medicine. First, existing tools often require advanced programming skills, creating a barrier for clinical researchers \cite{jarrett2023}. Second, data processing workflows lack standardization. Real-world clinical data is highly heterogeneous, often distributed in tables with inconsistent formats. Consequently, researchers must spend considerable time on tedious data cleaning and alignment prior to model training, which significantly slows down the research process. Finally, model implementation remains cumbersome. Advanced algorithms are often difficult to invoke directly, impeding the deployment of new technologies in real patient analysis \cite{makowski2021}.

To overcome these challenges and bridge the gap between algorithmic development and clinical application, we propose mtslearn, a highly integrated analysis framework explicitly designed for time-series data. The core contributions of mtslearn are twofold:

\begin{itemize}
    \item \textbf{Standardization of medical time-series data formats:} We establish a unified data interface specification that addresses the inconsistencies in raw data formats, so that data from multiple sources in different formats can be parsed and aligned automatically.
    \item \textbf{Construction of an accessible, full-pipeline framework:} The framework integrates a complete pipeline from raw data cleaning and feature engineering to model training and automated evaluation. With our highly encapsulated interfaces, clinicians can set up a complete analysis workflow in fewer than 20 lines of code. This allows them to focus on medical research rather than software implementation.
\end{itemize}

\section*{DESIGN AND DEVELOPMENT}

\subsection*{Problem Formulation}
\begin{figure}[htbp] 
    \centering 
    \includegraphics[width=\textwidth]{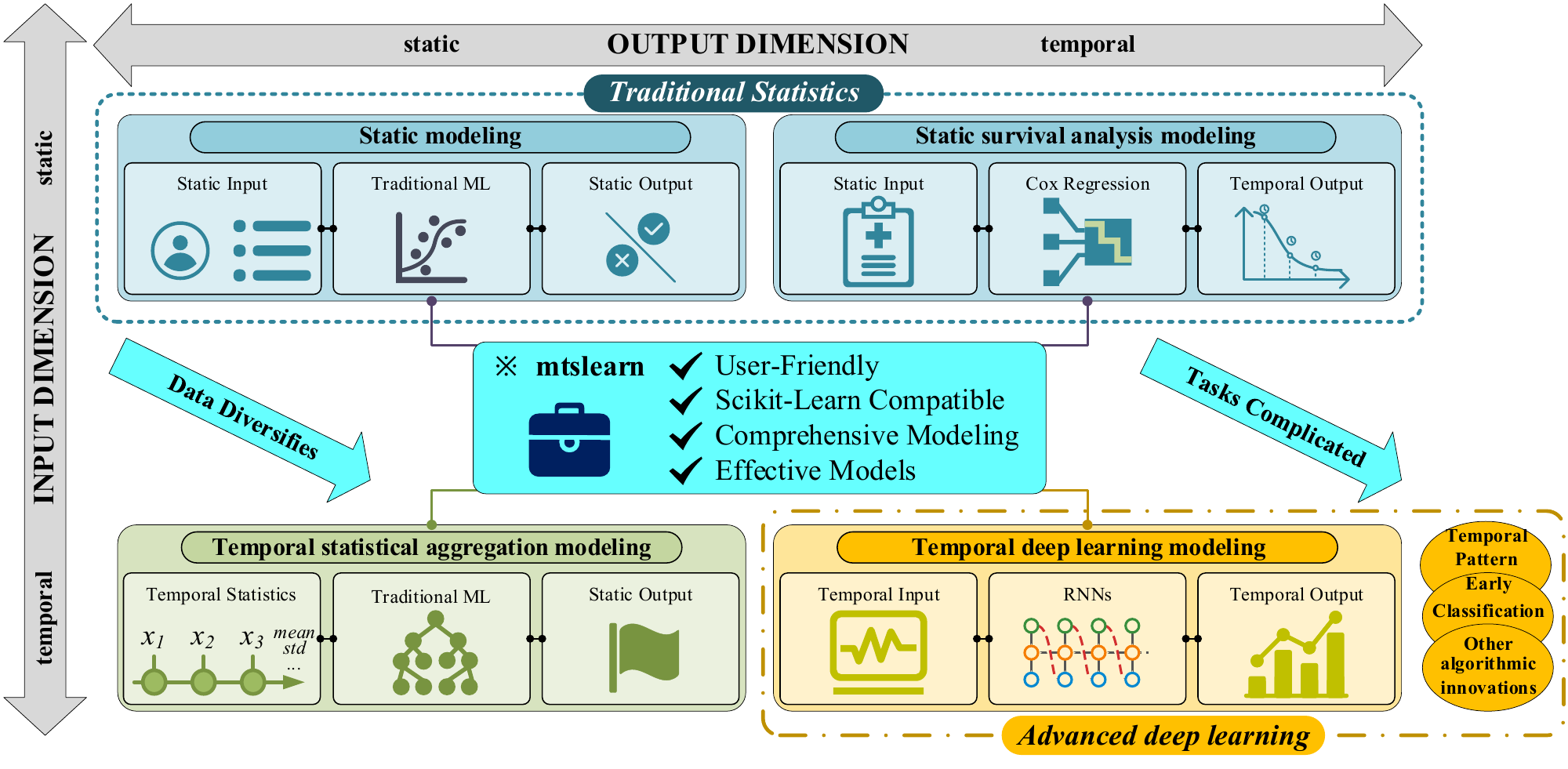} 
    \caption{Evolutionary paradigm of clinical prediction modeling. Rather than introducing complex deep learning algorithms for time-series data, this study builds upon traditional statistical methods and classic deep learning algorithms. We integrated established AI medical tools into a unified framework to simplify workflows for clinicians.}
    \label{fig:four_dim} 
\end{figure}

Clinical prediction models have evolved significantly recently driven by the growing availability of longitudinal data and the need for timely decisions. This evolutionary path from simple statistical mapping to complex time series modeling is shown in Figure \ref{fig:four_dim}.

Early models relied on structured tabular data, mapping a fixed set of patient variables to a single outcome. Logistic regression is a classic example and remains widely used due to its interpretability and robustness \cite{miyazaki2024,deng2022,hong2023,li2024}. Survival analysis extended this approach by making the prediction target time-dependent. For instance, the Cox proportional hazards model estimates risk over time while keeping the input features static, allowing researchers to address a broader range of clinical questions \cite{germer2024,lee2024,wen2024,zhang2024}.

The widespread use of electronic health records now provides a richer, longitudinal view of patient health. Initially, researchers handled repeated measurements by aggregating them into summary statistics, which forced temporal data into conventional tabular formats \cite{sun2022,wang2025}. More recently, deep learning sequence models have allowed researchers to model patient trajectories directly, capturing true temporal dependencies  \cite{hu2024,kuruwitaa.2025}. In clinical settings, these methods are further adapted. For example, T-LSTM \cite{baytas2017} can handle irregular sampling and varying time intervals, allowing models to better reflect real-world data collection processes.

Nowadays, researchers have multiple complementary ways to model patient data, but these methods are typically confined to isolated workflows. The wide variance in medical data formats, coupled with the steep engineering complexity of deep learning pipelines, severely limits the bedside adoption of advanced models. To overcome these barriers, rather than introducing increasingly complex deep learning algorithms, this study capitalizes on mature, established AI medical tools and traditional statistical methods. By integrating these well-validated tools into a unified, end-to-end framework—which cohesively handles static inputs, temporal data, and diverse targets—we eliminate redundant engineering steps and drastically simplify the modeling workflow for clinicians. Ultimately, this integration effectively bridges the gap between AI research and practical clinical application.

\subsection*{Data Specifications}
\begin{figure}[htbp] 
    \centering 
    \includegraphics[width=\textwidth]{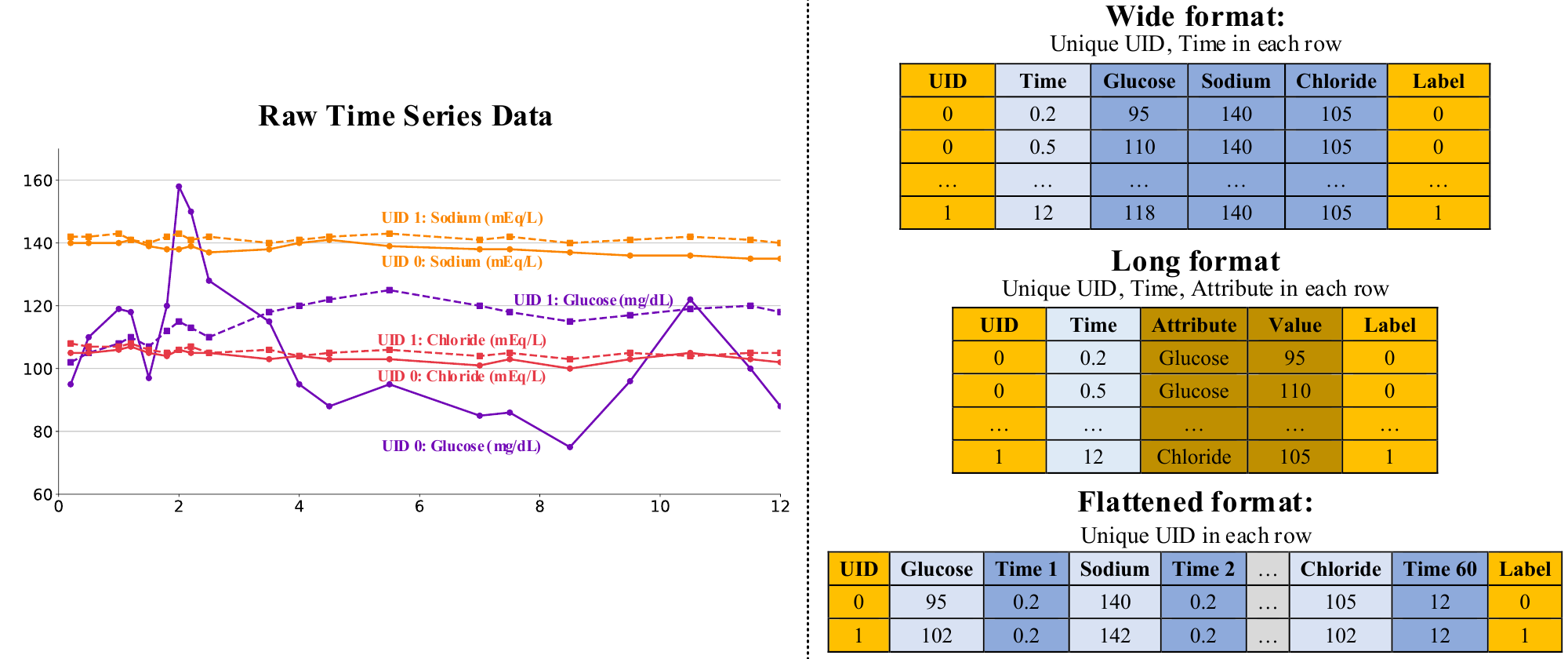} 
    \caption{Structural representations of heterogeneous clinical data. The schematic illustrates the three most common tabular formats encountered in clinical records: Wide format, Long format, and Flattened format. mtslearn standardizes these diverse inputs into a unified representation to significantly reduce manual preprocessing overhead.}
    \label{fig:different_tables} 
\end{figure}

To address the inconsistencies in clinical time series data and lower the technical barrier for clinicians, we first formalize the data input structures. As illustrated in Figure \ref{fig:different_tables}, we standardize three common tabular data schemas: wide, long, and flattened formats \cite{wickham2014}.

\begin{itemize}
    \item \textbf{Wide format}: This structure represents temporal snapshots, where each row corresponds to a specific time point for an individual subject, and feature measurements are organized into distinct columns.
    \item \textbf{Long format}: Highly suitable for sparse clinical data, this schema records each measurement as a separate row containing the identifier, timestamp, variable name, and corresponding value.
    \item \textbf{Flattened format}: This layout expands the data horizontally, using the subject ID as the unique row index. Measurement dimensions including time or specific features serve as column headers, encapsulating a patient's entire trajectory within a single row.
\end{itemize}

Based on the above format definition, mtslearn can easily integrate various datasets without requiring extensive manual preprocessing. Taking the common wide format as an example, clinicians simply specify basic metadata, including the patient ID \texttt{id\_col}, event timestamp \texttt{time\_col}, and outcome label \texttt{label\_col}. The framework then automatically parses and standardizes the underlying data structures, substantially reducing preprocessing overhead.

\subsection*{Design Rationale of mtslearn}
Despite the abundance of existing machine learning tools, the lack of end-to-end integration results in highly fragmented workflows. Complex operations and steep technical barriers deter clinicians from using these tools. This hinders the widespread adoption of advanced algorithms in clinical settings. Table \ref{tab:comparison} provides a detailed functional comparison of existing toolkits.

\begin{table}[htbp] 
    \centering
    \begin{threeparttable}
        \caption{Functional comparison of existing machine learning toolkits versus mtslearn. The table highlights capabilities across data preprocessing and model prediction, which shows that most existing toolkits focus on specific stages. ("End-to-End Pipeline" denotes a unified workflow integrating the entire lifecycle, from raw data processing to model evaluation.)}
        \label{tab:comparison}
        \begin{tabular}{l c c c c c c}
            \hline
            \multirow{2}{*}{\textbf{Name}} & \multicolumn{3}{c}{\textbf{Preprocessing}} & \multicolumn{2}{c}{\textbf{Predict}} & \multirow{2}{*}{\textbf{End-to-End}} \\ \cline{2-6}
            & \mhead{Data\\cleaning} & \mhead{Time-series\\resampling} & \mhead{Feature\\Stats} & \mhead{Static\\model} & \mhead{Time-series\\model} & \textbf{Pipelining} \\ \hline
            \textbf{TemporAI \cite{saveliev2023}} & $\checkmark$ & $\times$ & $\times$ & $\checkmark$ & $\checkmark$ & $\checkmark$ \\ \hline
            \textbf{Clairvoyance \cite{jarrett2023}} & $\checkmark$ & $\checkmark$ & $\times$ & $\checkmark$ & $\checkmark$ & $\checkmark$ \\ \hline
            \textbf{NeuroKit2 \cite{makowski2021}} & $\checkmark$ & $\checkmark$ & $\checkmark$ & $\times$ & $\times$ & $\times$ \\ \hline
            \textbf{Talent \cite{liu2024}} & $\checkmark$ & $\times$ & $\times$ & $\checkmark$ & $\times$ & $\checkmark$ \\ \hline
            \textbf{scikit-learn \cite{pedregosa2018}} & $\checkmark$ & $\times$ & $\times$ & $\checkmark$ & $\times$ & $\checkmark$ \\ \hline
            \textbf{scikit-survival \cite{JMLR:v21:20-729}} & $\times$ & $\times$ & $\times$ & $\checkmark$ & $\times$ & $\checkmark$ \\ \hline
            \textbf{mtslearn (ours)} & $\checkmark$ & $\checkmark$ & $\checkmark$ & $\checkmark$ & $\checkmark$ & $\checkmark$ \\ \hline
        \end{tabular}
        \begin{tablenotes}
            \item 
        \end{tablenotes}
    \end{threeparttable}
\end{table}

To address these limitations, the primary motivation behind mtslearn is to construct a highly integrated framework. Through standardized interfaces, it simultaneously supports both mainstream pathways for time-series data processing. By encapsulating tedious time-series preprocessing, feature engineering, and model training into accessible modules, mtslearn making it easier for clinical researchers to run machine learning experiments without deep programming knowledge.

\subsection*{End-to-End Pipeline Architecture}
\begin{figure}[htbp] 
    \centering 
    \includegraphics[width=0.95\textwidth]{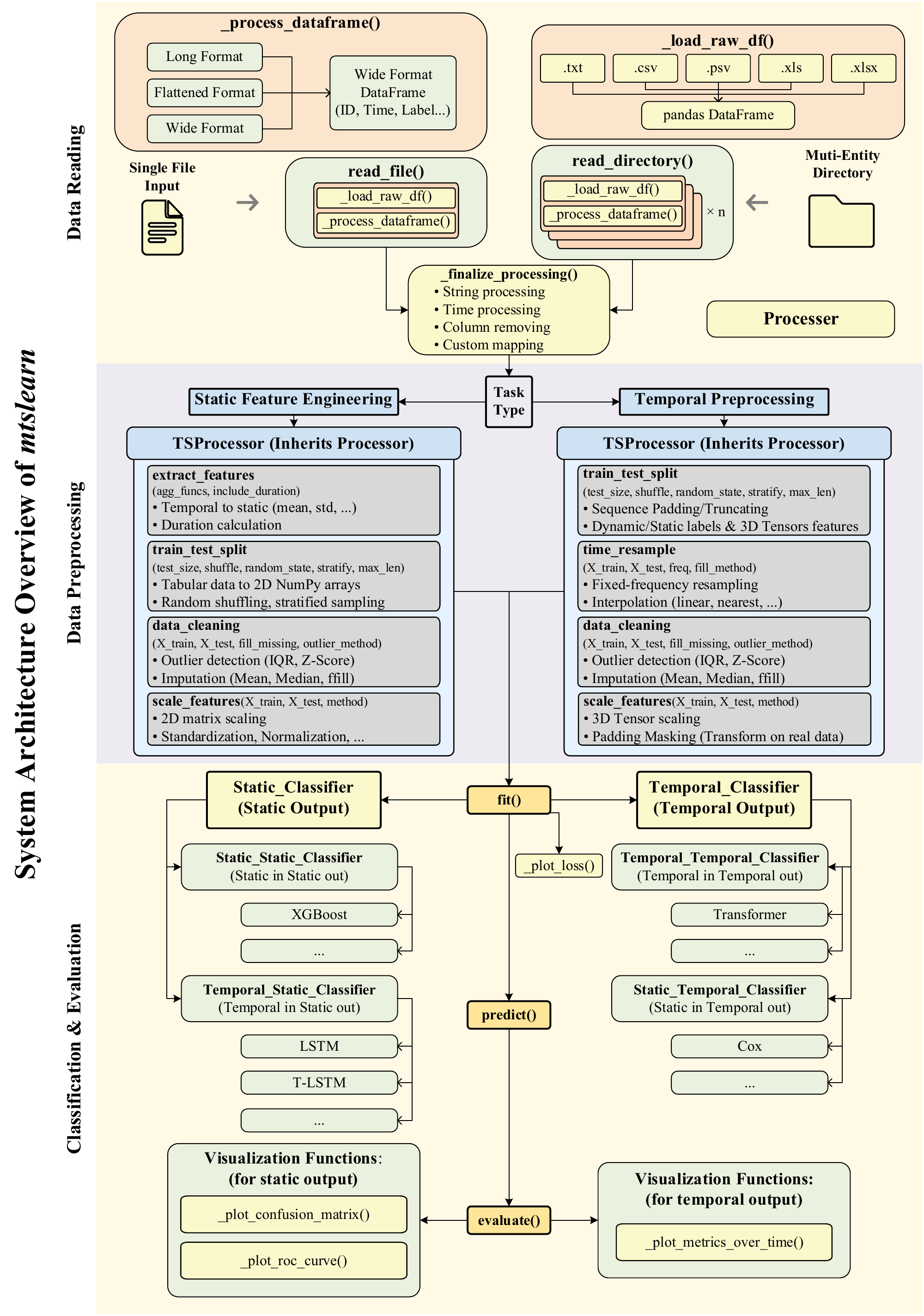} 
    \caption{Overview of the mtslearn pipeline.}
    \label{fig:overall} 
\end{figure}

mtslearn adopts a modular architecture to ensure maintainability and scalability. As illustrated in Figure \ref{fig:overall}, the core workflow begins with raw data reading, which then branches into static and time-series processing pipelines. Finally, the processed data are fed into the classifier module for model training, cross-validation, and comprehensive performance evaluation.

\subsection*{Core Functional Modules}
The framework contains four core modules. Together, they cover the entire analytical pipeline from data ingestion to model evaluation. The specific processing logic and capabilities of each module are detailed below.

\subsubsection*{Data Ingestion and Standardization: \texttt{Processor}}
The \texttt{Processor} module manages data ingestion and standardizes heterogeneous tabular data. It parses both individual files and multi-entity directories. The module transforms diverse input structures (long, wide, or flattened) into a unified wide representation, indexed by entity identifier and timestamp. It subsequently handles string parsing, time formatting, and the removal of redundant columns. Finally, it separates predictors from ground-truth labels before exporting the data.

\subsubsection*{Static Feature Engineering: \texttt{StaticProcessor}}
The \texttt{StaticProcessor} prepares temporal records for static machine learning algorithms. It achieves this by aggregating time-series sequences into static features. Specifically, it computes statistics like mean and variance, and extracts a duration metric to measure the observation window length. The module also handles standard data preparation tasks. These include dataset splitting, outlier removal, missing value imputation, and feature scaling.

\subsubsection*{Time-Series Data Processing: \texttt{TSProcessor}}
Deep learning models take fixed-size 3D arrays as input. \texttt{TSProcessor} converts variable-length multivariate sequences into this format by padding with trailing zeros and interpolating irregular time steps. For missing values, it first tries to fill from other observations of the same subject; if none are available, it falls back to global statistics. For missing data, the module uses a localized fallback imputation method. Missing values are first imputed using other observations from the same subject; if none exist, global statistics are used instead. During feature scaling, masking mechanisms are applied so that padded zeros do not distort the calculations.

\subsubsection*{Modeling and Evaluation Paradigms: \texttt{Static\_Classifier} and \texttt{Temporal\_Classifier}}
The modeling pipeline is encapsulated by two foundational classes, \texttt{Static\_Classifier} and \texttt{Temporal\_Classifier}, which handle static and continuous outputs, respectively. Together, they offer a unified interface for training optimization and tracking loss. Depending on the input and output data structures, the framework supports four different algorithmic paradigms.

First, for tasks predicting single-point outcomes from static baselines, the toolkit integrates models like XGBoost \cite{chen2016}. Second, for sequential inputs with static outcomes, it supports standard LSTMs \cite{vennerod2021} and Time-Aware LSTMs \cite{baytas2017}. The latter is specifically designed for irregular recording intervals common in electronic health records. Third, when predicting static baselines into longitudinal or time-to-event outcomes, the framework incorporates models such as the Cox proportional hazards model \cite{cox1972}. Finally, it integrates models like the Transformer \cite{vaswani2023} to map input sequences directly to continuous output sequences.

After training, these classes generate performance diagnostics based on the specific output type. Static outcome models produce confusion matrices and ROC curves. For models with continuous temporal outputs, the framework generates a line graph that tracks metrics including Precision, Recall, F1 score and ROC-AUC relative to the temporal sequence.

\subsection*{Framework Extensibility and Algorithm Integration}

Beyond classic built-in algorithms, mtslearn features an open architecture that allows users to easily integrate custom models. This design pattern enables researchers to integrate cutting-edge algorithms into the existing framework without modifying the source code. To ensure compatibility, custom model classes need only adhere to the framework's interface protocol by implementing the standard \texttt{\_\_init\_\_}, \texttt{fit}, and \texttt{predict} methods. Listing \ref{box:example0} fully illustrates how a custom model such as ”Random Forest“ can be added to mtslearn:

\begin{mybox}{box:example0}{Example of dynamic algorithm integration.}
    \begin{lstlisting}[language=Python]
from sklearn.ensemble import RandomForestClassifier
clf = Static_Static_Classifier(model_type='RandomForest')
clf.MODELS['RandomForest'] = RandomForestClassifier
clf.DEFAULT_CONFIGS['RandomForest'] = {'n_estimators': 100, 'max_depth': 5}
clf.fit(X_train_static, y_train_static)
clf.evaluate(X_test_static, y_test_static)
    \end{lstlisting}
\end{mybox}

This design grants advanced users substantial flexibility, ensuring that the framework can continuously evolve alongside rapid advancements in the machine learning community.

\subsection*{Installation of mtslearn}
The mtslearn framework is distributed via the Python Package Index (PyPI) and can be installed using the command \lstinline|pip install mtslearn|. The core architecture is built upon matplotlib, numpy, pandas, scikit-learn, pytorch, xgboost and lifelines.

\section*{VALIDATION AND EVALUATION}

This section evaluates the practical utility of the mtslearn toolkit through end-to-end clinical workflows. By implementing experiments across medical time-series prediction tasks, we demonstrate how the toolkit streamlines the critical stages of data preprocessing, feature engineering, and model evaluation. To illustrate the core functional capabilities of the framework, the following demonstrations focus on two representative paradigms, namely feature-aggregated static prediction and end-to-end time-series prediction.

\subsection*{COVID-19 Mortality Classification}

First, we conducted experiments on the COVID19-375 dataset, which represents a typical task with time-series inputs and a static categorical output aimed at predicting the final patient outcomes. For this task, we applied a static-input, static-output modeling paradigm by first aggregating the raw temporal sequences into static features. Utilizing the `StaticProcessor` module, we extracted aggregate statistical features including mean, std, max and min from multidimensional time series, while also incorporating temporal duration information. Subsequently, the workflow automatically handled missing values via mean imputation, addressed outliers using the interquartile range method, and standardized the features. Finally, an XGBoost classification model was trained and evaluated using the `Static\_Static\_Classifier` module. The code and the final prediction results are shown in Listing \ref{box:example1}.

\begin{mybox}{box:example1}{Usage example on COVID19-375}
    \begin{lstlisting}[language=Python]
from mtslearn import StaticProcessor, Static_Static_Classifier

data_config = {
    "file_path": 'data/covid19_data/375_patients_example.xlsx',
    "data_type": 'wide',        "time_col": 'RE_DATE',
    "id_col": 'PATIENT_ID',     "label_col": 'outcome',
}

static_processor = StaticProcessor()
static_processor.read_file(**data_config)
static_processor.extract_features(agg_funcs=['mean', 'std', 'max', 'min', 'median'], include_duration=True)
X_train_static, X_test_static, y_train_static, y_test_static = static_processor.train_test_split(test_size=0.3, shuffle=True, random_state=42, stratify=True)
X_train_static, X_test_static = static_processor.data_cleaning(X_train_static, X_test_static, fill_missing='mean', outlier_method='iqr')
X_train_static, X_test_static = static_processor.scale_features(X_train_static, X_test_static, method='standardize')

model = Static_Static_Classifier(model_type='XGB')
model.fit(X_train_static, y_train_static)
model.evaluate(X_test_static, y_test_static)
    \end{lstlisting}
    \includegraphics[width=0.47\textwidth]{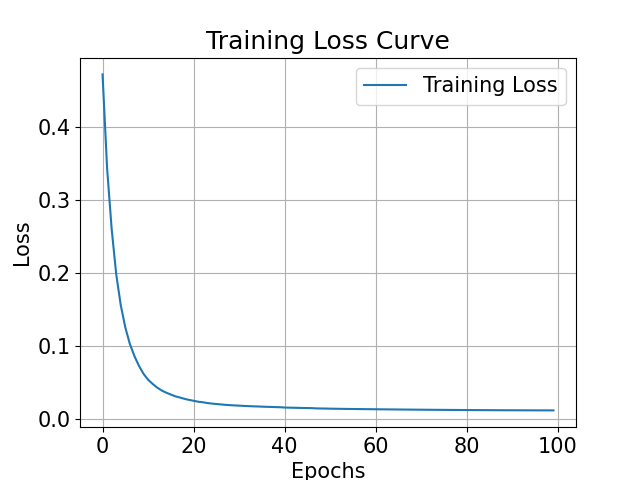}
    \begin{verbatim}
--- XGB Classification Report ---
              precision    recall  f1-score   support

           0       0.97      0.97      0.97        61
           1       0.96      0.96      0.96        52

    accuracy                           0.96       113
   macro avg       0.96      0.96      0.96       113
weighted avg       0.96      0.96      0.96       113
    \end{verbatim}
    \includegraphics[width=0.5\textwidth]{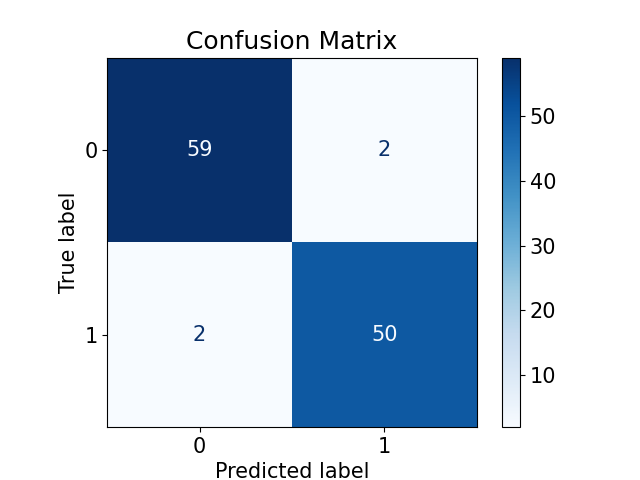}
    \includegraphics[width=0.5\textwidth]{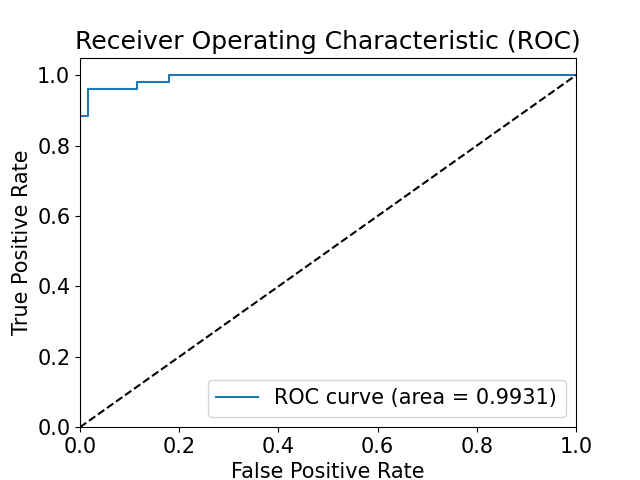}
\end{mybox}

\subsection*{Sepsis Early Diagnosis}

\begin{figure}[htbp] 
    \centering 
    \includegraphics[width=0.8\textwidth]{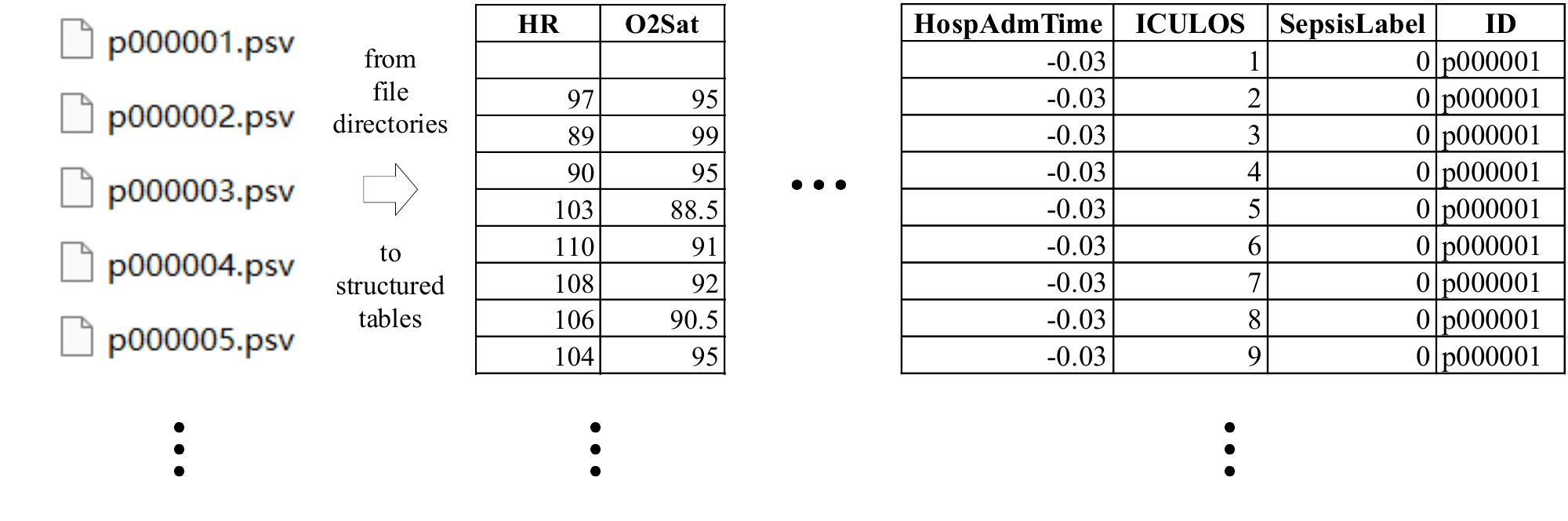} 
    \caption{Before and after data reading.} 
    \label{fig:table2} 
\end{figure}

Second, we carried out a sepsis prediction task on the PhysioNet 2019 dataset. This dataset belongs to a more complex category involving time-series inputs mapped to time-series categorical outputs. To address this task, we adopted deep learning methods that directly process temporal sequences. Using the TSProcessor module, we read and organized data from directories, truncated sequences to a maximum length of 30, and split the dataset with stratification. During the data cleaning phase, a forward-fill strategy was applied to handle temporal missing values, followed by outlier correction and data standardization. Lastly, by invoking the `Temporal\_Temporal\_Classifier` module, we trained and evaluated a time-series classification model based on the Transformer architecture. Figure \ref{fig:table2} visually illustrates the transformation process before and after data reading: multiple separate files are read, formatted, and aligned, and consolidated into a standardized wide format. The complete process code and performance metrics for this task are presented in Listing \ref{box:example2}.

\begin{mybox}{box:example2}{Usage example on PhysioNet 2019}
    \begin{lstlisting}[language=Python]
from mtslearn import TSProcessor, Temporal_Temporal_Classifier

data_config = {"dir_path": 'data/sepsis_data/set_A/', 
    "data_type": 'wide',        "time_col": 'ICULOS', 
    "label_col": 'SepsisLabel', "label_type": 'temporal'}
ts_processor = TSProcessor()
ts_processor.read_directory(**data_config)

X_train, X_test, y_train, y_test = ts_processor.train_test_split(test_size=0.2, shuffle=True, random_state=42, max_len=30)
X_train, X_test = ts_processor.data_cleaning(X_train, X_test, fill_missing='ffill', outlier_method='iqr')
X_train, X_test = ts_processor.scale_features(X_train, X_test, method='standardize')

classifier = Temporal_Temporal_Classifier(model_type='Transformer')
classifier.fit(X_train, y_train)
classifier.evaluate(X_test, y_test)
    \end{lstlisting}
    \includegraphics[width=0.7\textwidth]{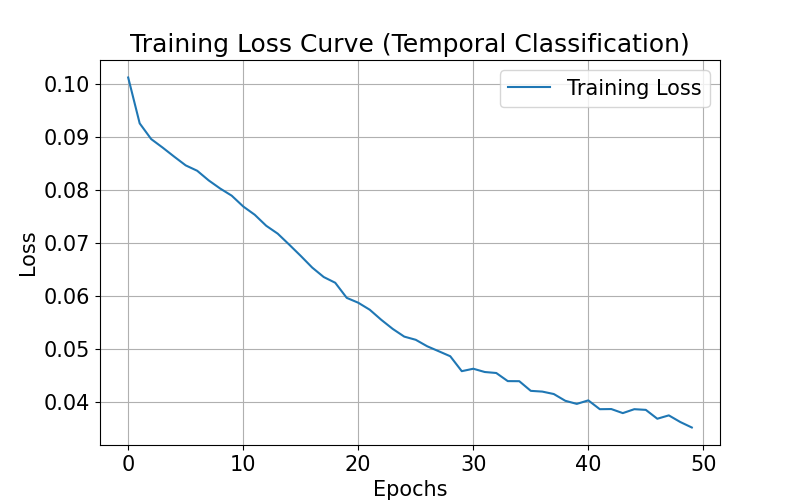}
    \begin{verbatim}
--- Transformer Temporal Classification Report (flattened) ---
              precision    recall  f1-score   support

         0.0       0.98      0.99      0.98    118629
         1.0       0.41      0.24      0.30      3411

    accuracy                           0.97    122040
   macro avg       0.70      0.61      0.64    122040
weighted avg       0.96      0.97      0.97    122040
    \end{verbatim}
    \includegraphics[width=0.75\textwidth]{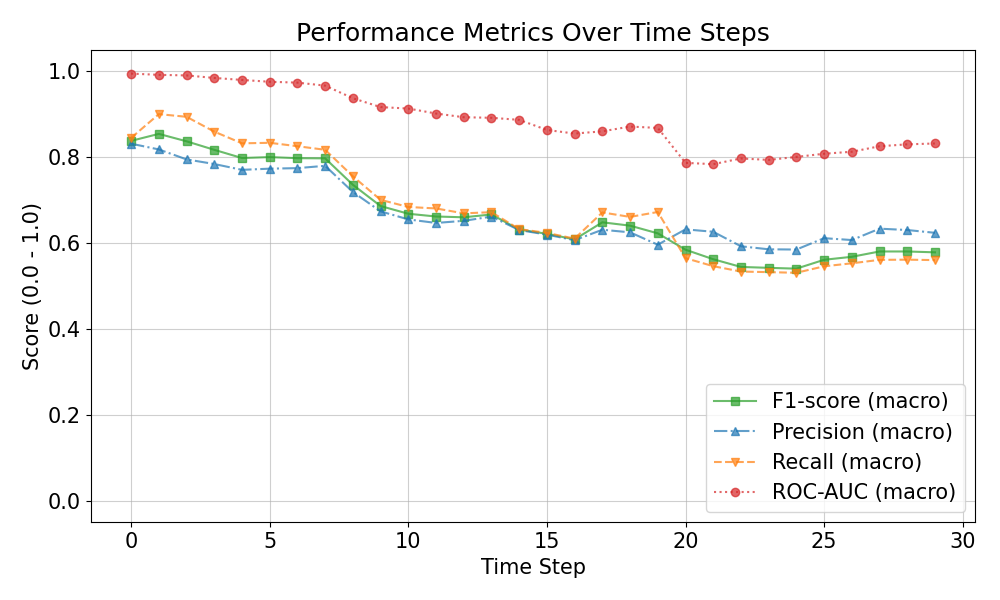}
\end{mybox}

\section*{DISCUSSION AND CONCLUSION}

To help close the technical gap between clinical medicine and machine learning, we developed mtslearn, a toolkit specifically for medical time-series analysis. mtslearn allows clinicians with limited programming experience to manage data processing and modeling tasks independently. This makes building clinical decision-support models from real-world data faster and easier.

The main advantage of mtslearn is its simplicity and ease of use. It follows the scikit-learn API, making it easy to learn. The toolkit is comprehensive, covering data loading and preprocessing while providing suitable modeling approaches for four common input-output types in clinical tasks. Essentially, mtslearn takes a step further than traditional statistical methods.

Despite these advantages, the current work has a few limitations. First, while mtslearn already includes some basic deep learning capabilities, it has not yet expanded to more advanced deep learning architectures. This limits its full potential when dealing with highly complex tasks. Second, the current system lacks built-in explainability. Without explicit clinical reasoning, it is difficult to show how the models make decisions, which is a common barrier to earning doctors' trust.

In future work, we plan to introduce explainable AI modules to provide clinical justifications for model decisions, thereby increasing physician trust in these systems. Concurrently, we will continuously integrate more advanced deep learning models. This will ensure the toolkit remains capable of addressing increasingly complex clinical analysis requirements.

\section*{RESOURCE AVAILABILITY}

mtslearn is publicly available at \url{https://github.com/PKUDigitalHealth/mtslearn}, and \url{https://pypi.org/project/mtslearn/}.

\section*{ACKNOWLEDGMENTS}

This work was supported by the National Natural Science Foundation of China (62102008), CCF-Tencent Rhino-Bird Open Research Fund (CCF-Tencent RAGR20250108), CCF-Zhipu Large Model Innovation Fund (CCF-Zhipu202414), PKU-OPPO Fund (BO202301, BO202503), Research Project of Peking University in the State Key Laboratory of Vascular Homeostasis and Remodeling (2025-SKLVHR-YCTS-02), Beijing Municipal Science and Technology Commission (Z251100000725008).

\section*{DECLARATION OF INTERESTS}

Shenda Hong is an associate editor of Health Data Science. He was not involved in the review or decision-making process for this manuscript. The other authors declare no competing interests.

\bibliography{references}

@inproceedings{baytas2017,
  title = {Patient {{Subtyping}} via {{Time-Aware LSTM Networks}}},
  booktitle = {Proceedings of the 23rd {{ACM SIGKDD International Conference}} on {{Knowledge Discovery}} and {{Data Mining}}},
  author = {Baytas, Inci M. and Xiao, Cao and Zhang, Xi and Wang, Fei and Jain, Anil K. and Zhou, Jiayu},
  year = 2017,
  month = aug,
  pages = {65--74},
  publisher = {ACM},
  address = {Halifax NS Canada},
  doi = {10.1145/3097983.3097997},
  urldate = {2026-03-08},
  isbn = {978-1-4503-4887-4},
  langid = {english}
}

@inproceedings{chen2016,
  title = {{{XGBoost}}: {{A Scalable Tree Boosting System}}},
  shorttitle = {{{XGBoost}}},
  booktitle = {Proceedings of the 22nd {{ACM SIGKDD International Conference}} on {{Knowledge Discovery}} and {{Data Mining}}},
  author = {Chen, Tianqi and Guestrin, Carlos},
  year = 2016,
  month = aug,
  series = {{{KDD}} '16},
  pages = {785--794},
  publisher = {Association for Computing Machinery},
  address = {New York, NY, USA},
  doi = {10.1145/2939672.2939785},
  urldate = {2026-03-08},
  isbn = {978-1-4503-4232-2}
}

@article{deng2022,
  title = {{[Long short-term memory and Logistic regression for mortality risk prediction of intensive care unit patients with stroke]}},
  author = {Deng, Y. H. and Jiang, Y. and Wang, Z. Y. and Liu, S. and Wang, Y. X. and Liu, B. H.},
  year = 2022,
  month = jun,
  journal = {Beijing Da Xue Xue Bao. Yi Xue Ban = Journal of Peking University. Health Sciences},
  volume = {54},
  number = {3},
  pages = {458--467},
  issn = {1671-167X},
  doi = {10.19723/j.issn.1671-167X.2022.03.010},
  langid = {chi},
  pmcid = {PMC9197695},
  pmid = {35701122}
}

@article{germer2024,
  title = {Survival Analysis for Lung Cancer Patients: {{A}} Comparison of {{Cox}} Regression and Machine Learning Models},
  shorttitle = {Survival Analysis for Lung Cancer Patients},
  author = {Germer, Sebastian and Rudolph, Christiane and Labohm, Louisa and Katalinic, Alexander and Rath, Natalie and Rausch, Katharina and Holleczek, Bernd and {AI-CARE Working Group} and Handels, Heinz},
  year = 2024,
  month = nov,
  journal = {International Journal of Medical Informatics},
  volume = {191},
  pages = {105607},
  issn = {1872-8243},
  doi = {10.1016/j.ijmedinf.2024.105607},
  langid = {english},
  pmid = {39208536}
}

@article{hu2024,
  title = {A Prediction Approach to {{COVID-19}} Time Series with {{LSTM}} Integrated Attention Mechanism and Transfer Learning},
  author = {Hu, Bin and Han, Yaohui and Zhang, Wenhui and Zhang, Qingyang and Gu, Wen and Bi, Jun and Chen, Bi and Xiao, Lishun},
  year = 2024,
  month = dec,
  journal = {BMC Medical Research Methodology},
  volume = {24},
  number = {1},
  pages = {323},
  issn = {1471-2288},
  doi = {10.1186/s12874-024-02433-w},
  urldate = {2026-03-09},
  langid = {english}
}

@misc{jarrett2023,
  title = {Clairvoyance: {{A Pipeline Toolkit}} for {{Medical Time Series}}},
  shorttitle = {Clairvoyance},
  author = {Jarrett, Daniel and Yoon, Jinsung and Bica, Ioana and Qian, Zhaozhi and Ercole, Ari and van der Schaar, Mihaela},
  year = 2023,
  month = oct,
  number = {arXiv:2310.18688},
  eprint = {2310.18688},
  primaryclass = {cs},
  publisher = {arXiv},
  doi = {10.48550/arXiv.2310.18688},
  urldate = {2026-03-08},
  archiveprefix = {arXiv}
}

@article{kuruwitaa.2025,
  title = {A {{Hybrid Bidirectional Deep Learning Model Using HRV}} for {{Prediction}} of {{ICU Mortality Risk}} in {{TBI Patients}}},
  author = {Kuruwita A., Hasitha and Ng, Shu Kay and Liew, Alan Wee-Chung and Ross, Kelvin and Richards, Brent and Kumar, Kuldeep and Haseler, Luke and Zhang, Ping},
  year = 2025,
  month = dec,
  journal = {Journal of Healthcare Informatics Research},
  volume = {9},
  number = {4},
  pages = {629--655},
  issn = {2509-4971, 2509-498X},
  doi = {10.1007/s41666-025-00209-5},
  urldate = {2026-03-09},
  langid = {english}
}

@article{lee2024,
  title = {Personalized Prediction of Survival Rate with Combination of Penalized {{Cox}} Models in Patients with Colorectal Cancer},
  author = {Lee, Seon Hwa and Cha, Jae Myung and Shin, Seung Jun},
  year = 2024,
  month = jun,
  journal = {Medicine},
  volume = {103},
  number = {24},
  pages = {e38584},
  issn = {1536-5964},
  doi = {10.1097/MD.0000000000038584},
  langid = {english},
  pmcid = {PMC11175897},
  pmid = {38875378}
}

@misc{liu2024,
  title = {{{TALENT}}: {{A Tabular Analytics}} and {{Learning Toolbox}}},
  shorttitle = {{{TALENT}}},
  author = {Liu, Si-Yang and Cai, Hao-Run and Zhou, Qi-Le and Ye, Han-Jia},
  year = 2024,
  month = jul,
  number = {arXiv:2407.04057},
  eprint = {2407.04057},
  primaryclass = {cs},
  publisher = {arXiv},
  doi = {10.48550/arXiv.2407.04057},
  urldate = {2026-03-08},
  archiveprefix = {arXiv}
}

@article{luo2024,
  title = {Recent {{Advances}} in {{Wearable Healthcare Devices}}: {{From Material}} to {{Application}}},
  shorttitle = {Recent {{Advances}} in {{Wearable Healthcare Devices}}},
  author = {Luo, Xiao and Tan, Handong and Wen, Weijia},
  year = 2024,
  month = apr,
  journal = {Bioengineering},
  volume = {11},
  number = {4},
  pages = {358},
  issn = {2306-5354},
  doi = {10.3390/bioengineering11040358},
  urldate = {2026-03-08},
  langid = {english}
}

@article{makowski2021,
  title = {{{NeuroKit2}}: {{A Python}} Toolbox for Neurophysiological Signal Processing},
  shorttitle = {{{NeuroKit2}}},
  author = {Makowski, Dominique and Pham, Tam and Lau, Zen J. and Brammer, Jan C. and Lespinasse, Fran{\c c}ois and Pham, Hung and Sch{\"o}lzel, Christopher and Chen, S. H. Annabel},
  year = 2021,
  month = aug,
  journal = {Behavior Research Methods},
  volume = {53},
  number = {4},
  pages = {1689--1696},
  issn = {1554-3528},
  doi = {10.3758/s13428-020-01516-y},
  urldate = {2026-03-08},
  langid = {english}
}

@article{miyazaki2024,
  title = {Logistic Regression Analysis and Machine Learning for Predicting Post-Stroke Gait Independence: A Retrospective Study},
  shorttitle = {Logistic Regression Analysis and Machine Learning for Predicting Post-Stroke Gait Independence},
  author = {Miyazaki, Yuta and Kawakami, Michiyuki and Kondo, Kunitsugu and Hirabe, Akiko and Kamimoto, Takayuki and Akimoto, Tomonori and Hijikata, Nanako and Tsujikawa, Masahiro and Honaga, Kaoru and Suzuki, Kanjiro and Tsuji, Tetsuya},
  year = 2024,
  month = sep,
  journal = {Scientific Reports},
  volume = {14},
  number = {1},
  pages = {21273},
  publisher = {Nature Publishing Group},
  issn = {2045-2322},
  doi = {10.1038/s41598-024-72206-4},
  urldate = {2026-03-09},
  copyright = {2024 The Author(s)},
  langid = {english}
}

@misc{saveliev2023,
  title = {{{TemporAI}}: {{Facilitating Machine Learning Innovation}} in {{Time Domain Tasks}} for {{Medicine}}},
  shorttitle = {{{TemporAI}}},
  author = {Saveliev, Evgeny S. and van der Schaar, Mihaela},
  year = 2023,
  month = jan,
  number = {arXiv:2301.12260},
  eprint = {2301.12260},
  primaryclass = {cs},
  publisher = {arXiv},
  doi = {10.48550/arXiv.2301.12260},
  urldate = {2026-03-08},
  archiveprefix = {arXiv}
}

@article{shen2025,
  title = {Twenty-{{Five Years}} of {{Evolution}} and {{Hurdles}} in {{Electronic Health Records}} and {{Interoperability}} in {{Medical Research}}: {{Comprehensive Review}}},
  shorttitle = {Twenty-{{Five Years}} of {{Evolution}} and {{Hurdles}} in {{Electronic Health Records}} and {{Interoperability}} in {{Medical Research}}},
  author = {Shen, Yun and Yu, Jiamin and Zhou, Jian and Hu, Gang},
  year = 2025,
  month = jan,
  journal = {Journal of Medical Internet Research},
  volume = {27},
  pages = {e59024},
  issn = {1438-8871},
  doi = {10.2196/59024},
  urldate = {2026-03-08},
  langid = {english}
}

@article{sun2022,
  title = {A {{Machine Learning Pipeline}} for {{Mortality Prediction}} in the {{ICU}}},
  author = {Sun, Yang and Zhou, Yi-Hui},
  year = 2022,
  journal = {International Journal of Digital Health},
  volume = {2},
  number = {1},
  pages = {3},
  issn = {2634-4580},
  doi = {10.29337/ijdh.44},
  urldate = {2026-03-09},
  langid = {american}
}

@misc{vennerod2021,
  title = {Long {{Short-term Memory RNN}}},
  author = {Venner{\o}d, Christian Bakke and Kj{\ae}rran, Adrian and Bugge, Erling Stray},
  year = 2021,
  month = may,
  number = {arXiv:2105.06756},
  eprint = {2105.06756},
  primaryclass = {cs},
  publisher = {arXiv},
  doi = {10.48550/arXiv.2105.06756},
  urldate = {2026-03-08},
  archiveprefix = {arXiv}
}

@article{wang2025,
  title = {Performance {{Comparison}} of {{Models}} for {{Short-Term Mortality Prediction}} in {{ICU Patients}}},
  author = {Wang, Junhe},
  year = 2025,
  month = oct,
  journal = {Theoretical and Natural Science},
  volume = {132},
  pages = {85--91},
  issn = {2753-8826, 2753-8818},
  doi = {10.54254/2753-8818/2025.DL27346},
  urldate = {2026-03-09},
  copyright = {Copyright (c) 2025 Junhe Wang},
  langid = {english}
}

@article{wickham2014,
  title = {Tidy {{Data}}},
  author = {Wickham, Hadley},
  year = 2014,
  journal = {Journal of Statistical Software},
  volume = {59},
  number = {10},
  issn = {1548-7660},
  doi = {10.18637/jss.v059.i10},
  urldate = {2026-03-08},
  langid = {english}
}

@misc{pedregosa2018,
  title = {Scikit-Learn: {{Machine Learning}} in {{Python}}},
  shorttitle = {Scikit-Learn},
  author = {Pedregosa, Fabian and Varoquaux, Ga{\"e}l and Gramfort, Alexandre and Michel, Vincent and Thirion, Bertrand and Grisel, Olivier and Blondel, Mathieu and M{\"u}ller, Andreas and Nothman, Joel and Louppe, Gilles and Prettenhofer, Peter and Weiss, Ron and Dubourg, Vincent and Vanderplas, Jake and Passos, Alexandre and Cournapeau, David and Brucher, Matthieu and Perrot, Matthieu and Duchesnay, {\'E}douard},
  year = 2018,
  month = jun,
  number = {arXiv:1201.0490},
  eprint = {1201.0490},
  primaryclass = {cs},
  publisher = {arXiv},
  doi = {10.48550/arXiv.1201.0490},
  urldate = {2026-03-09},
  archiveprefix = {arXiv}
}

@article{JMLR:v21:20-729,
  title = {Scikit-Survival: A Library for Time-to-Event Analysis Built on Top of Scikit-Learn},
  author = {P{\"o}lsterl, Sebastian},
  year = 2020,
  journal = {Journal of Machine Learning Research},
  volume = {21},
  number = {212},
  pages = {1--6}
}

@article{niu2025,
  title = {Modelling {{Patient Longitudinal Data}} for {{Clinical Decision Support}}: {{A Case Study}} on {{Emerging AI Healthcare Technologies}}},
  shorttitle = {Modelling {{Patient Longitudinal Data}} for {{Clinical Decision Support}}},
  author = {Niu, Shuai and Ma, Jing and Yin, Qing and Wang, Zhihua and Bai, Liang and Yang, Xian},
  year = 2025,
  month = apr,
  journal = {Information Systems Frontiers},
  volume = {27},
  number = {2},
  pages = {409--427},
  issn = {1572-9419},
  doi = {10.1007/s10796-024-10513-x},
  urldate = {2026-03-09},
  langid = {english}
}

@article{hong2023,
  title = {{{simpleNomo}}: {{A Python Package}} of {{Making Nomograms}} for {{Visualizable Calculation}} of {{Logistic Regression Models}}},
  shorttitle = {{{simpleNomo}}},
  author = {Hong, Haoyang and Hong, Shenda},
  year = 2023,
  month = jun,
  journal = {Health Data Science},
  volume = {3},
  pages = {0023},
  publisher = {American Association for the Advancement of Science},
  doi = {10.34133/hds.0023},
  urldate = {2026-03-20}
}

@article{li2024,
  title = {Federated {{Learning}} in {{Healthcare}}: {{A Benchmark Comparison}} of {{Engineering}} and {{Statistical Approaches}} for {{Structured Data Analysis}}},
  shorttitle = {Federated {{Learning}} in {{Healthcare}}},
  author = {Li, Siqi and Miao, Di and Wu, Qiming and Hong, Chuan and D'Agostino, Danny and Li, Xin and Ning, Yilin and Shang, Yuqing and Wang, Ziwen and Liu, Molei and Fu, Huazhu and Ong, Marcus Eng Hock and Haddadi, Hamed and Liu, Nan},
  year = 2024,
  month = dec,
  journal = {Health Data Science},
  volume = {4},
  pages = {0196},
  publisher = {American Association for the Advancement of Science},
  doi = {10.34133/hds.0196},
  urldate = {2026-03-20}
}

@article{wen2024,
  title = {Survival {{Disparities}} among {{Cancer Patients Based}} on {{Mobility Patterns}}: {{A Population-Based Study}}},
  shorttitle = {Survival {{Disparities}} among {{Cancer Patients Based}} on {{Mobility Patterns}}},
  author = {Wen, Fengyu and Zhang, Yike and Yang, Chao and Li, Pengfei and Wang, Qing and Zhang, Luxia},
  year = 2024,
  month = nov,
  journal = {Health Data Science},
  volume = {4},
  pages = {0198},
  publisher = {American Association for the Advancement of Science},
  doi = {10.34133/hds.0198},
  urldate = {2026-03-20}
}

@article{zhang2024,
  title = {Association of {{Smoking}} with {{Chronic Kidney Disease Stages}} 3 to 5: {{A Mendelian Randomization Study}}},
  shorttitle = {Association of {{Smoking}} with {{Chronic Kidney Disease Stages}} 3 to 5},
  author = {Zhang, Zhilong and Zhang, Feifei and Zhang, Xiaomeng and Lu, Lanlan and Zhang, Luxia},
  year = 2024,
  month = nov,
  journal = {Health Data Science},
  volume = {4},
  pages = {0199},
  publisher = {American Association for the Advancement of Science},
  doi = {10.34133/hds.0199},
  urldate = {2026-03-20}
}

@article{cox1972,
  title = {Regression {{Models}} and {{Life-Tables}}},
  author = {Cox, D. R.},
  year = 1972,
  month = jan,
  journal = {Journal of the Royal Statistical Society: Series B (Methodological)},
  volume = {34},
  number = {2},
  pages = {187--202},
  issn = {0035-9246},
  doi = {10.1111/j.2517-6161.1972.tb00899.x},
  urldate = {2026-03-22}
}

@misc{vaswani2023,
  title = {Attention {{Is All You Need}}},
  author = {Vaswani, Ashish and Shazeer, Noam and Parmar, Niki and Uszkoreit, Jakob and Jones, Llion and Gomez, Aidan N. and Kaiser, Lukasz and Polosukhin, Illia},
  year = 2023,
  month = aug,
  number = {arXiv:1706.03762},
  eprint = {1706.03762},
  primaryclass = {cs},
  publisher = {arXiv},
  doi = {10.48550/arXiv.1706.03762},
  urldate = {2026-03-22},
  archiveprefix = {arXiv}
}

\end{document}